\documentclass[11pt,a4paper]{article}
\usepackage[hyperref]{eacl2021}
\usepackage{times}
\usepackage{latexsym}

\usepackage{times}
\usepackage{url}
\usepackage{latexsym}
\usepackage{danudefs}
\usepackage{multirow}

\usepackage{microtype}

\aclfinalcopy 

\usepackage[english]{babel}
\usepackage{hyperref}
\addto\captionsenglish{%
}
\addto\extrasenglish{%
}

\usepackage{esvect}
\usepackage{graphicx}
\usepackage{comment}
\usepackage{caption}
\usepackage{subcaption}
\usepackage{booktabs}

\usepackage{color}

\usepackage{todonotes}

\makeatletter
\if@todonotes@disabled

\else

\fi
\makeatother

\usepackage{booktabs}
\usepackage{graphics}
\usepackage{autobreak}
\usepackage{adjustbox}
\usepackage{breqn}

\title{Dictionary-based Debiasing of Pre-trained Word Embeddings}

\author{
    Masahiro Kaneko\\
    Tokyo Metropolitan University \\
  {\tt kaneko-masahiro@ed.tmu.ac.jp}
    \And
    Danushka Bollegala\Thanks{ Danushka Bollegala holds concurrent appointments as a Professor at University of Liverpool and as an Amazon Scholar. This paper describes work performed at the University of Liverpool and is not associated with Amazon.} \\
  University of Liverpool, Amazon\\
  {\tt danushka@liverpool.ac.uk}}

\date{}

\begin{document}
\maketitle
\begin{abstract}
  Word embeddings trained on large corpora have shown to encode high levels of unfair discriminatory gender, racial, religious and ethnic biases.
  In contrast, human-written dictionaries describe the meanings of words in a concise, objective and an unbiased manner.
  We propose a method for debiasing pre-trained word embeddings using dictionaries, without requiring access to the original training resources or any knowledge regarding the word embedding algorithms used. 
  Unlike prior work, our proposed method does not require the types of biases to be pre-defined in the form of word lists, and learns the constraints that must be satisfied by unbiased word embeddings automatically from dictionary definitions of the words.
  Specifically, we learn an encoder to generate a debiased version of an input word embedding such that it
  (a) retains the semantics of the pre-trained word embeddings,
  (b) agrees with the unbiased definition of the word according to the dictionary, and
  (c) remains orthogonal to the vector space spanned by any biased basis vectors in the pre-trained word embedding space.
  Experimental results on standard benchmark datasets  show that the proposed method can accurately remove unfair biases encoded in pre-trained word embeddings, while preserving useful semantics.
\end{abstract}

\section{Introduction}

Despite pre-trained word embeddings are useful due to their low dimensionality, memory and compute efficiency, they have shown to encode not only the semantics of words but also unfair discriminatory biases such as gender, racial or religious biases~\cite{Tolga:NIPS:2016,Zhao:2018aa,Rudinger:2018aa,Zhao:2018ab,Elazar:EMNLP:2018,kaneko-bollegala-2019-gender}.
On the other hand, human-written dictionaries act as an impartial, objective and unbiased source of word meaning.
Although methods that learn word embeddings by purely using dictionaries have been proposed~\cite{tissier-gravier-habrard:2017:EMNLP2017}, they have coverage and data sparseness related issues because pre-compiled dictionaries do not capture the meanings of neologisms or provide numerous contexts as in a corpus. 
Consequently, prior work has shown that word embeddings learnt from large text corpora to outperform those created from dictionaries in downstream NLP tasks~\cite{Alsuhaibani:AKBC:2019,Bollegala:AAAI:2016}.

We must overcome several challenges when using dictionaries to debias pre-trained word embeddings. 
First, not all words in the embeddings will appear in the given dictionary.
Dictionaries often have limited coverage and will not cover neologisms, orthographic variants of words etc. that are likely to appear in large corpora.
A lexicalised debiasing method would generalise poorly to the words not in the dictionary.
Second, it is not known apriori what biases are hidden inside a set of pre-trained word embedding vectors.
Depending on the source of documents used for training the embeddings, different types of biases will be learnt and amplified by different word embedding learning algorithms to different degrees~\cite{zhao-EtAl:2017:EMNLP20173}.

Prior work on debiasing required that the biases to be pre-defined~\cite{kaneko-bollegala-2019-gender}.
For example, Hard-Debias~\cite[HD;][]{Tolga:NIPS:2016} and Gender Neutral Glove~\cite[GN-GloVe;][]{Zhao:2018ab} require lists of \emph{male} and \emph{female} pronouns for defining the \emph{gender} direction. 
However, gender bias is only one of the many biases that exist in pre-trained word embeddings.
It is inconvenient to prepare lists of words covering all different types of biases we must remove from pre-trained word embeddings.
Moreover, such pre-compiled word lists are likely to be incomplete and inadequately cover some biases.
Indeed, \newcite{gonen-goldberg-2019-lipstick} showed empirical evidence that such debiasing methods do not remove all discriminative biases from word embeddings.
Unfair biases have adversely affected several NLP tasks such as machine translation~\cite{Vanmassenhove:2018} and language generation~\cite{sheng-etal-2019-woman}.
Racial biases have also been shown to affect criminal prosecutions~\cite{manzini-etal-2019-black} and career adverts~\cite{Lambrecht_2016}.
These findings show the difficulty of defining different biases using pre-compiled lists of words, which is a requirement in previously proposed debiasing methods for static word embeddings.

We propose a method that uses a dictionary as a source of bias-free definitions of words for debiasing pre-trained word embeddings\footnote{Code and debiased embeddings: \url{https://github.com/kanekomasahiro/dict-debias}}.
Specifically, we learn an encoder that filters-out biases from the input embeddings.
The debiased embeddings are required to simultaneously satisfy three criteria:
(a) must preserve all non-discriminatory information in the pre-trained embeddings (\emph{semantic preservation}),
(b) must be similar to the dictionary definition of the words (\emph{dictionary agreement}), and
(c) must be orthogonal to the subspace spanned by the basis vectors in the pre-trained word embedding space that corresponds to discriminatory biases (\emph{bias orthogonality}).
We implement the semantic preservation and dictionary agreement using two decoders, whereas the bias orthogonality is enforced by a parameter-free projection.
The debiasing encoder and the decoders are learnt end-to-end by a joint optimisation method.
Our proposed method is agnostic to the details of the algorithms used to learn the input word embeddings.
Moreover, unlike counterfactual data augmentation methods for debiasing~\cite{Zmigrod:2019,hall-maudslay-etal-2019-name}, we do \emph{not} require access to the original training resources used for learning the input word embeddings.

Our proposed method overcomes the above-described challenges as follows.
First, instead of learning a lexicalised debiasing model, we operate on the word embedding space when learning the encoder.
Therefore, we can use the words that are in the intersection of the vocabularies of the pre-trained word embeddings and the dictionary to learn the encoder, enabling us to generalise  to the words not in the dictionary.
Second, we do \emph{not} require pre-compiled word lists specifying the biases.
The dictionary acts as a clean, unbiased source of word meaning that can be considered as \emph{positive} examples of debiased meanings.
In contrast to the existing debiasing methods that require us to pre-define \emph{what to remove}, the proposed method can be seen as using the dictionary as a guideline for \emph{what to retain} during debiasing.

We evaluate the proposed method using four standard benchmark datasets for evaluating the biases in word embeddings: 
Word Embedding Association Test~\cite[WEAT;][]{Caliskan2017SemanticsDA}, 
Word Association Test~\cite[WAT;][]{du-etal-2019-exploring}, 
SemBias~\cite{Zhao:2018ab} 
and WinoBias~\cite{Zhao:2018aa}. 
Our experimental results show that the proposed debiasing method accurately removes unfair biases from three widely used pre-trained embeddings: Word2Vec~\cite{Mikolov:NAACL:2013}, GloVe~\cite{Glove} and fastText~\cite{bojanowski2017enriching}.
Moreover, our evaluations on semantic similarity and word analogy benchmarks show that the proposed debiasing method preserves useful semantic information in word embeddings, while removing unfair biases.

\section{Related Work}
\label{sec:related}

Dictionaries have been popularly used for learning word embeddings~\cite{Budanitsky,Budanitsky:NAACL:2001,Jiang:1997}.
Methods that use both dictionaries (or lexicons) and corpora to jointly learn word embeddings~\cite{tissier-gravier-habrard:2017:EMNLP2017,Alsuhaibani:AKBC:2019,Bollegala:AAAI:2016} or post-process~\cite{Glavas:2018aa,faruqui-EtAl:2015:NAACL-HLT} have also been proposed.
However, learning embeddings from dictionaries alone results in coverage and data sparseness issues~\cite{,Bollegala:AAAI:2016} and does not guarantee bias-free embeddings~\cite{Lauscher2019AreWC}. 
To the best of our knowledge, we are the first to use dictionaries for debiasing pre-trained word embeddings.

\newcite{Tolga:NIPS:2016} proposed a post-processing approach that projects gender-neutral words into a subspace, which is orthogonal to the gender dimension defined by a list of gender-definitional words.
They refer to words associated with gender (e.g., \emph{she}, \emph{actor}) as gender-definitional words, and the remainder gender-neutral.
They proposed a \emph{hard-debiasing} method where the gender direction is computed as the vector difference between the embeddings of the corresponding gender-definitional words, and a \emph{soft-debiasing} method, which balances the objective of preserving the inner-products between the original word embeddings, while projecting the word embeddings into a subspace orthogonal to the gender definitional words. 
Both hard and soft debiasing methods ignore gender-definitional words during the subsequent debiasing process, and focus only on words that are \emph{not} predicted as gender-definitional by the classifier. 
Therefore, if the classifier erroneously predicts a stereotypical word as a gender-definitional word, it would not get debiased.

\newcite{Zhao:2018ab} modified the GloVe~\cite{Glove} objective to learn gender-neutral word embeddings (GN-GloVe) from a given corpus.
They maximise the squared $\ell_{2}$ distance between gender-related sub-vectors, while simultaneously minimising the GloVe objective.
Unlike, the above-mentioned methods, \newcite{kaneko-bollegala-2019-gender} proposed a post-processing method to preserve gender-related information with autoencoder~\cite{kaneko-bollegala-2020-autoencoding}, while removing discriminatory biases from stereotypical cases (GP-GloVe).
However, all prior debiasing methods require us to pre-define the biases in the form of explicit word lists containing gender and stereotypical word associations.
In contrast we use dictionaries as a source of bias-free semantic definitions of words and do not require pre-defining the biases to be removed.
Although we focus on static word embeddings in this paper, unfair biases have been found in contextualised word embeddings as well~\cite{Zhao:2019a,Vig:2019,bordia-bowman-2019-identifying,may-etal-2019-measuring}.

Adversarial learning methods~\cite{Xie:NIPS:2017,Elazar:EMNLP:2018,Li:2018ab} for debiasing first encode the inputs and then two classifiers are jointly trained -- one predicting the target task (for which we must ensure high prediction accuracy) and the other protected attributes (that must not be easily predictable). 
However,  \newcite{Elazar:EMNLP:2018} showed that although it is possible to obtain chance-level development-set accuracy for the protected attributes during training, a post-hoc classifier trained on the encoded inputs can still manage to reach substantially high accuracies for the protected attributes. 
They conclude that adversarial learning alone does not guarantee invariant representations for the protected attributes.
\newcite{Ravfogel:2020} found that iteratively projecting word embeddings to the null space of the gender direction to further improve the debiasing performance. 

To evaluate biases, \newcite{Caliskan2017SemanticsDA} proposed the Word Embedding Association Test (WEAT) inspired by the Implicit Association Test~\cite[IAT;][]{IAT}.
\newcite{ethayarajh-etal-2019-understanding} showed that WEAT to be systematically overestimating biases and proposed a correction.
The ability to correctly answer gender-related word analogies~\cite{Zhao:2018ab} and resolve gender-related coreferences~\cite{Zhao:2018aa,Rudinger:2018aa} have been used as extrinsic tasks for evaluating the bias in word embeddings.
We describe these evaluation benchmarks later in \autoref{sec:datasets}.

\section{Dictionary-based Debiasing}

Let us denote the $n$-dimensional pre-trained word embedding of a word $w$ by $\vec{w} \in \R^n$ trained on some resource $\cC$ such as a text corpus.
Moreover, let us assume that we are given a dictionary $\cD$ containing the definition, $s(w)$ of  $w$. 
If the pre-trained embeddings distinguish among the different senses of $w$, then we can use the gloss for the corresponding sense of $w$ in the dictionary as $s(w)$.
However, the majority of word embedding learning methods do not produce sense-specific word embeddings. 
In this case, we can either use all glosses for $w$ in $\cD$ by concatenating or select the gloss for the dominant (most frequent) sense of $w$\footnote{Prior work on debiasing static word embeddings do not use contextual information that is required for determining word senses. Therefore, for comparability reasons we do neither.}.
Without any loss of generality, in the remainder of this paper, we will use $s(w)$ to collectively denote a gloss selected by any one of the above-mentioned criteria with or without considering the word senses (in \autoref{sec:gloss}, we evaluate the effect of using all vs. dominant gloss).

Next, we define the objective functions optimised by the proposed method for the purpose of learning unbiased word embeddings.
Given, $\vec{w}$, we model the debiasing process as the task of learning an encoder, $E(\vec{w}; \vec{\theta}_e)$ that returns an $m (\leq n)$-dimensional debiased version of $\vec{w}$. 
In the case where we would like to preserve the dimensionality of the input embeddings, we can set $m = n$, or $m < n$ to further compress the debiased embeddings.


Because the pre-trained embeddings encode rich semantic information from a large text corpora, often far exceeding the meanings covered in the dictionary, we must preserve this semantic information as much as possible during the debiasing process. We refer to this constraint as \emph{semantic preservation}.
Semantic preservation is likely to lead to good performance in downstream NLP applications that use pre-trained word embeddings.
For this purpose, we decode the encoded version of $\vec{w}$ using a decoder, $D_{c}$, parametrised by $\vec{\theta}_{c}$ and define $J_{c}$ to be the reconstruction loss given by \eqref{eq:Jc}.
\begin{align}
 J_{c}(w) = \norm{\vec{w} - D_{c}(E(\vec{w}; \vec{\theta}_{e}); \vec{\theta}_{c})}_{2}^{2}
 \label{eq:Jc}
\end{align}

Following our assumption that the dictionary definition, $s(w)$, of $w$ is a concise and unbiased description of the meaning of $w$, we would like to ensure that the encoded version of $\vec{w}$ is similar to $s(w)$. 
We refer to this constraint as \emph{dictionary agreement}.
To formalise dictionary agreement empirically, we first represent $s(w)$ by a sentence embedding vector $\vec{s}(w) \in \R^{n}$.
Different sentence embedding methods can be used for this purpose such as convolutional neural networks~\cite{kim:2014:EMNLP2014}, recurrent neural networks~\cite{Elmo} or transformers~\cite{BERT}.
For the simplicity, we use the smoothed inverse frequency~\cite[SIF;][]{Arora:ICLR:2017} for creating $\vec{s}(w)$ in this paper.
SIF computes the embedding of a sentence as the weighted average of the pre-trained word embeddings of the words in the sentence, where the weights are computed as the inverse unigram probability. 
Next, the first principal component vector of the sentence embeddings are removed.
The dimensionality of the sentence embeddings created using SIF is equal to that of the pre-trained word embeddings used.
Therefore, in our case we have both $\vec{w}, \vec{s}(w) \in \R^{n}$.

We decode the debiased embedding $E(\vec{w}; \vec{\theta}_{e})$ of $w$ using a decoder $D_{d}$, parametrised by $\vec{\theta}_{d}$ and compute the squared $\ell_{2}$ distance between it and $\vec{s}(w)$ to define an objective $J_{d}$ given by \eqref{eq:Jd}.
\begin{align}
 \label{eq:Jd}
 J_{d}(w) = \norm{\vec{s}(w) - D_{d}(E(\vec{w}; \vec{\theta}_{e}); \vec{\theta}_{d})}_{2}^{2}
\end{align}

Recalling that our goal is to remove unfair biases from pre-trained word embeddings and we assume dictionary definitions to be free of such biases, we define an objective function that explicitly models this requirement. 
We refer to this requirement as the \emph{bias orthogonality} of the debiased embeddings.
For this purpose, we first project the pre-trained word embedding $\vec{w}$ of a word $w$ into a subspace that is orthogonal to the dictionary definition vector $\vec{s}(w)$.
Let us denote this projection by $\phi(\vec{w}, \vec{s}(w)) \in \R^{n}$. 
We require that the debiased word embedding, $E(\vec{w}; \vec{\theta}_{e})$, must be orthogonal to $\phi(\vec{w}, \vec{s}(w))$, and formalise this as the minimisation of the squared inner-product given in \eqref{eq:Ja}.
\begin{align}
 \label{eq:Ja}
 J_{a}(w) = \left(E(\phi(\vec{w}, \vec{s}(w)); \vec{\theta}_{e})\T E(\vec{w}; \vec{\theta}_{e})\right)^{2}
\end{align}
Note that because $\phi(\vec{w}, \vec{s}(w))$ lives in the space spanned by the original (prior to encoding) vector space, we must first encode it using $E$ before considering the orthogonality requirement. 

To derive $\phi(\vec{w}, \vec{s}(w))$, let us assume the $n$-dimensional basis vectors in the $R^{n}$ vector space spanned by the pre-trained word embeddings to be $\vec{b}_{1}, \vec{b}_{2}, \ldots, \vec{b}_{n}$.
Moreover, without loss of generality,  let the subspace spanned by the subset of the first $k (<n)$ basis vectors $\vec{b}_{1}, \vec{b}_{2}, \ldots, \vec{b}_{k}$ to be $\cB \subseteq \R^{n}$.
The projection $\vec{v}_{\cB}$ of a vector $\vec{v} \in \R^{n}$ onto $\cB$ can be expressed using the basis vectors as in \eqref{eq:proj}.
\begin{align}
\label{eq:proj}
\vec{v}_{\cB} = \sum_{j=1}^{k} (\vec{v}\T\vec{b}_{j}) \vec{b}_{j}
\end{align}
To show that $\vec{v} - \vec{v}_{\cB}$ is orthogonal to $\vec{v}_{\cB}$ for any $\vec{v} \in \cB$, let us express $\vec{v} - \vec{v}_{\cB}$ using the basis vectors as given in \eqref{eq:b1}.
\begin{align}
\vec{v} - \vec{v}_{\cB} &=  \sum_{i=1}^{n} (\vec{v}\T\vec{b}_{i}) \vec{b}_{i} - \sum_{j=1}^{k} (\vec{v}\T\vec{b}_{j}) \vec{b}_{j} \nonumber \\
                        &= \sum_{i=k+1}^{n} (\vec{v}\T\vec{b}_{i}) \vec{b}_{i}
\label{eq:b1}
\end{align}
We see that there are no basis vectors in common between the summations in  \eqref{eq:proj} and \eqref{eq:b1}.
Therefore, $\vec{v}_{\cB} \T (\vec{v} - \vec{v}_{\cB}) = 0$ for $\forall \vec{v} \in \cB$.
  
Considering that $\vec{s}(w)$ defines a direction that does not contain any unfair biases, we can compute the vector rejection of $\vec{w}$ on $\vec{s}(w)$ following this result.
Specifically, we subtract the projection of $\vec{w}$ along the unit vector defining the direction of $\vec{s}(w)$ to compute $\phi$ as in \eqref{eq:phi}.
\begin{align}
\label{eq:phi}
\phi(\vec{w}, \vec{s}(w)) = \vec{w} - \vec{w}\T \vec{s}(w) \frac{\vec{s}(w)}{\norm{\vec{s}(w)}}
\end{align}
We consider the linearly-weighted sum of the above-defined three objective functions as the total objective function as given in \eqref{eq:total}.
\begin{align}
 \label{eq:total}
 J(w) = \alpha J_{c}(w) + \beta J_{d}(w) + \gamma J_{a}(w)
\end{align}
Here, $\alpha, \beta, \gamma \geq 0$ are scalar coefficients satisfying $\alpha + \beta + \gamma = 1$.
Later, in \autoref{sec:exp} we experimentally determine the values of $\alpha, \beta$ and $\gamma$ using a development dataset.

\section{Experiments}
\label{sec:exp}

\subsection{Word Embeddings}

In our experiments, we use the following publicly available pre-trained word embeddings:  \textbf{Word2Vec}\footnote{\url{https://code.google.com/archive/p/word2vec/}} (300-dimensional embeddings for ca. 3M words learned from Google News corpus~\cite{NIPS2013_5021}), 
\textbf{GloVe}\footnote{\url{https://github.com/stanfordnlp/GloVe}} (300-dimensional embeddings for ca. 2.1M words learned from the Common Crawl~\cite{Glove}), and 
\textbf{fastText}\footnote{\url{https://fasttext.cc/docs/en/english-vectors.html}} (300-dimensional embeddings for ca. 1M words learned from Wikipedia 2017, UMBC webbase corpus and statmt.org news~\cite{bojanowski2017enriching}).

As the dictionary definitions, we used the glosses in the WordNet~\cite{fellbaum98wordnet}, which has been popularly used to learn word embeddings in prior work ~\cite{tissier-gravier-habrard:2017:EMNLP2017,Bosc2018AutoEncodingDD,Washio2019BridgingTD}.
However, we note that our proposed method does not depend on any WordNet-specific features, thus in principle can be applied to any dictionary containing definition sentences.
Words that do not appear in the vocabulary of the pre-trained embeddings are ignored when computing $\vec{s}(w)$ for the headwords $w$ in the dictionary.
Therefore, if all the words in a dictionary definition are ignored, then the we remove the corresponding headwords from training.
Consequently, we are left with 54,528, 64,779 and 58,015 words respectively for \textbf{Word2Vec}, \textbf{GloVe} and \textbf{fastText} embeddings in the training dataset.
We randomly sampled 1,000 words from this dataset and held-out as a development set for the purpose of tuning various hyperparameters in the proposed method.

$E$, $D_{c}$ and $D_{d}$ are implemented as single-layer feed forward neural networks with a hyperbolic tangent activation at the outputs.
It is known that pre-training is effective when using autoencoders $E$ and $D_{c}$ for debiasing~\cite{kaneko-bollegala-2019-gender}.
Therefore, we randomly select 5000 words from each pre-trained word embedding set and pre-train the autoencoders on those words with a mini-batch of size 512.
In pre-training, the model with the lowest loss according to \eqref{eq:Jc} in the development set for pre-traininng is selected.

\subsection{Hyperparameters}

During optimisation, we used dropout~\cite{Srivastava2014DropoutAS} with probability $0.05$ to $\vec{w}$ and $E(w)$.
We used Adam~\cite{Kingma:ICLR:2015} with initial learning rate set to $0.0002$ as the optimiser to find the parameters $\vec{\theta}_{e}, \vec{\theta}_{c}$, and $\vec{\theta}_{d}$ and a mini-batch size of 4.
The optimal values of all hyperparameters are found by minimising the total loss over the development dataset
following a Monte-Carlo search.
We found these optimal hyperparameter values of $\alpha = 0.99998$,  $\beta = 0.00001$ and $\gamma = 0.00001$.
Note that the scale of different losses are different and the absolute values of hyperparameters do \emph{not} indicate the significance of a component loss. 
For example, if we rescale all losses to the same range then we have 
$L_c = 0.005\alpha$, $L_d = 0.269\beta$ and $L_a = 21.1999\gamma$. 
Therefore, debiasing ($L_d$) and orthogonalisation  ($L_a$) contributions are significant. 

We utilized a GeForce GTX 1080 Ti.
The debiasing is completed in less than an hour because our method is only a fine-tuning technique.
The parameter size of our debiasing model is 270,900.

\subsection{Evaluation Datasets}
\label{sec:datasets}

We use the following datasets to evaluate the degree of the biases in word embeddings.
\paragraph{WEAT:} Word Embedding Association Test~\cite[WEAT;][]{Caliskan2017SemanticsDA}, quantifies various biases (e.g. gender, race and age) using semantic similarities between word embeddings.
It compares two same size sets of \emph{target} words $\cX$ and $\cY$ (e.g. European and African names), with two sets of \emph{attribute} words $\cA$ and $\cB$ (e.g. \emph{pleasant} vs. \emph{unpleasant}). 
The bias score, $s(\cX,\cY,\cA,\cB)$, for each target is calculated as follows:
\begin{align}
s(\cX,\cY,\cA,\cB) &= \sum_{\vec{x} \in \cX} k(\vec{x}, \cA, \cB) \nonumber \\
                   &- \sum_{\vec{y} \in \cY} k(\vec{y}, \cA, \cB) \\
k(\vec{t}, \cA, \cB) &= \textrm{mean}_{\vec{a} \in \cA} f(\vec{t}, \vec{a}) \nonumber \\
                     &- \textrm{mean}_{\vec{b} \in \cB} f(\vec{t}, \vec{b})
\end{align}
Here, $f$ is the cosine similarity between the word embeddings.
The one-sided $p$-value for the permutation test regarding $\cX$ and $\cY$ is calculated as the probability of $s(\cX_i,\cY_i,\cA,\cB) > s(\cX,\cY,\cA,\cB)$.
The effect size is calculated as the normalised measure given by \eqref{eq:effect}.
\begin{align}
\label{eq:effect}
\frac{\textrm{mean}_{x \in \cX} s(x, \cA,\cB) - \textrm{mean}_{y \in \cY} s(y, \cA, \cB)}{\textrm{sd}_{t \in \cX \cup \cY} s(t, \cA, \cB)}
\end{align}

\begin{table*}[t]
\centering
\small
\begin{tabular}{lcccccc}
\toprule
\multirow{2}{*}{Embeddings} & Word2Vec & GloVe & fastText \\
                                         & Org/Deb             & Org/Deb      & Org/Deb      \\ \midrule
T1: flowers vs. insects                        & $1.46^\dag$/$\mathbf{1.35}^\dag$ & $\mathbf{1.48}^\dag$/$1.54^\dag$          & $1.29^\dag$/$\mathbf{1.09}^\dag$ \\
T2: instruments vs. weapons                    & $1.56^\dag$/$\mathbf{1.43}^\dag$ & $1.49^\dag$/$\mathbf{1.41}^\dag$ & $1.56^\dag$/$\mathbf{1.34}^\dag$ \\
T3: European vs. African American names       & $0.46^\dag$/$\mathbf{0.16}^\dag$ & $1.33^\dag$/$\mathbf{1.04}^\dag$ & $0.79^\dag$/$\mathbf{0.46}^\dag$ \\
T4: male vs. female                           & $1.91^\dag$/$\mathbf{1.87}^\dag$ & $1.86^\dag$/$\mathbf{1.85}^\dag$ & $1.65^\dag$/$\mathbf{1.42}^\dag$ \\
T5: math vs. art                              & $0.85^\dag$/$\mathbf{0.53}^\dag$ & $\mathbf{0.43}^\dag$/$0.82^\dag$          & $1.14^\dag$/$\mathbf{0.86}^\dag$ \\
T6: science vs. art                           & $1.18^\dag$/$\mathbf{0.96}^\dag$ & $\mathbf{1.21}^\dag$/$1.44^\dag$          & $1.16^\dag$/$\mathbf{0.88}^\dag$ \\
T7: physical vs. mental conditions             & 0.90/$\mathbf{0.57}$      & 1.03/$\mathbf{0.98}$      & 0.83/$\mathbf{0.63}$ \\
T8: older vs. younger names                   & $\mathbf{-0.08}$/$-0.10$              & $1.07^\dag$/$\mathbf{0.92}^\dag$ & -0.32/$\mathbf{-0.13}$  \\ \midrule
T9: WAT                                      & $0.48^\dag$/$\mathbf{0.45}^\dag$ & $0.59^\dag$/$\mathbf{0.58}^\dag$ & $0.54^\dag$/$\mathbf{0.51}^\dag$ \\ \bottomrule
\end{tabular}
\caption{Rows T1-T8 show WEAT bias effects for the cosine similarity and row T9 shows the Pearson correlations on the WAT dataset with cosine similarity. $\dag$ indicates bias effects that are insignificant at $\al < 0.01$.}
\label{tbl:weat_wat}
\end{table*}

\paragraph{WAT:} Word Association Test (WAT) is a method to measure gender bias over a large set of words~\cite{du-etal-2019-exploring}.
It calculates the gender information vector for each word in a word association graph created with Small World of Words project~\cite[SWOWEN;][]{Deyne2019TheW} by propagating information related to masculine and feminine words $(w_m^i, w_f^i) \in \cL$ using a random walk approach~\cite{Zhou2003LearningWL}.
The gender information is represented as a 2-dimensional vector ($b_m$, $b_f$), where $b_m$ and $b_f$ denote respectively the masculine and feminine orientations of a word.
The gender information vectors of masculine words, feminine words and other words are initialised respectively with vectors (1, 0), (0, 1) and (0, 0).
The bias score of a word is defined as $\log(b_m / b_f)$.
We evaluate the gender bias of word embeddings using the Pearson correlation coefficient between the bias score of each word and 
the score given by \eqref{eq:bias-score} computed as the averaged difference of cosine similarities between masculine and feminine words.
\begin{align}
\label{eq:bias-score}
\frac{1}{|\cL|} \sum_{i=1}^{|\cL|} \left( f(w, w_m^i) - f(w, w_f^i) \right)
\end{align}

\paragraph{SemBias:} SemBias dataset~\cite{Zhao:2018ab} contains three types of word-pairs: 
(a) \textbf{Definition}, a gender-definition word pair (e.g. hero -- heroine), 
(b) \textbf{Stereotype}, a gender-stereotype word pair (e.g., manager -- secretary) 
and (c) \textbf{None}, two other word-pairs with similar meanings unrelated to gender (e.g., jazz -- blues, pencil -- pen). 
We use the cosine similarity between the $\vv{he} - \vv{she}$ gender directional vector and $\vec{a} - \vec{b}$ in above word pair $(a,b)$ lists to measure gender bias.
\newcite{Zhao:2018ab} used a subset of 40 instances associated with 2 seed word-pairs, not used in the training split, to evaluate the generalisability of a debiasing method.
For unbiased word embeddings, we expect high similarity scores in \textbf{Definition} category and low similarity scores in \textbf{Stereotype} and \textbf{None} categories.

\begin{table}[t]
\centering
\small
\begin{tabular}{lccc}
\toprule
\multirow{2}{*}{Embeddings} & Word2Vec 		& GloVe 						& fastText \\
                          			 & Org/Deb    		&  Org/Deb  					& Org/Deb   \\ \midrule
definition 	      			& 83.0/\textbf{83.9} 	& 83.0/\textbf{83.4}   			& 92.0/\textbf{93.2}\\
stereotype        			& 13.4/\textbf{12.3} 	& 12.0/\textbf{11.4}   			& 5.5/\textbf{4.3} \\
none              			& \textbf{3.6}/3.9     	& \textbf{5.0}/5.2       			& \textbf{2.5}/\textbf{2.5} \\ \midrule
sub-definition    			& 50.0/\textbf{57.5} 	& \textbf{67.5}/\textbf{67.5}   	& 82.5/\textbf{85.0}  \\
sub-stereotype    			& 40.0/\textbf{32.5} 	& \textbf{27.5}/\textbf{27.5}   	& 12.5/\textbf{10.0}  \\
sub-none          			& \textbf{10.0}/\textbf{10.0} & \textbf{5.0}/\textbf{5.0}    & \textbf{5.0}/\textbf{5.0} \\ \bottomrule
\end{tabular}
\caption{Prediction accuracies for gender relational analogies on SemBias.}
\label{tbl:sembias}
\end{table}

\paragraph{WinoBias/OntoNotes:} We use the WinoBias dataset~\cite{Zhao:2018aa} and OntoNotes~\cite{weischedel2013ontonotes} for coreference resolution to evaluate the effectiveness of our proposed debiasing method in a downstream task.
WinoBias contains two types of sentences that require linking gendered pronouns to either male or female stereotypical occupations.
In \textbf{Type 1}, co-reference decisions must be made using world knowledge about some given circumstances. 
However, in \textbf{Type 2}, these tests can be resolved using syntactic information and understanding of the pronoun.
It involves two conditions: the pro-stereotyped (\textbf{pro}) condition links pronouns to occupations dominated by the gender of the pronoun, and the anti-stereotyped (\textbf{anti}) condition links pronouns to occupations not dominated by the gender of the pronoun.
For a correctly debiased set of word embeddings, the difference between \textbf{pro} and \textbf{anti} is expected to be small.
We use the model proposed by \newcite{lee-etal-2017-end} and implemented in AllenNLP~\cite{Gardner2017AllenNLP} as the coreference resolution method.

We used a bias comparing code\footnote{\url{https://github.com/hljames/compare-embedding-bias}} to evaluate \textbf{WEAT} dataset.
Since the \textbf{WAT} code was not published, we contacted the authors to obtain the code and used it for evaluation.
We used the evaluation code from GP-GloVe\footnote{\url{https://github.com/kanekomasahiro/gp_debias}} to evaluate \textbf{SemBias} dataset.
We used AllenNLP\footnote{\url{https://github.com/allenai/allennlp}} to evaluate \textbf{WinoBias} and \textbf{OntoNotes} datasets.
We used \textit{evaluate\_word\_pairs} function and \textit{evaluate\_word\_analogies} in gensim\footnote{\url{https://github.com/RaRe-Technologies/gensim}} to evaluate \textbf{word embedding benchmarks}.

\section{Results}
\subsection{Overall Results}

We initialise the word embeddings of the model by original (\textbf{Org}) and debiased (\textbf{Deb}) word embeddings and compare the coreference resolution accuracy using F1 as the evaluation measure.

\begin{table}[t]
\small
\centering
\begin{tabular}{lcccccc}
\toprule
\multirow{2}{*}{Embeddings} & Word2Vec & GloVe & fastText \\
                            & Org/Deb      & Org/Deb        & Org/Deb       \\ \midrule
Type 1-pro                        & 70.1/69.4          & 70.8/69.5            & 70.1/69.7           \\
Type 1-anti                        & 49.9/50.5          & 50.9/52.1            & 52.0/51.6           \\
Avg                         & \textbf{60.0}/\textbf{60.0} & \textbf{60.9}/60.8            & \textbf{61.1}/60.7           \\
Diff                        & 20.2/\textbf{18.9} & 19.9/\textbf{17.4}   & \textbf{18.1}/\textbf{18.1}  \\ \midrule
Type 2-pro                        & 84.7/83.7          & 79.6/78.9            & 83.8/82.5           \\
Type 2-anti                        & 77.9/77.5          & 66.0/66.4            & 75.1/76.4           \\
Avg                         & \textbf{81.3}/80.6          & \textbf{72.8}/72.7            & \textbf{79.5}/\textbf{79.5}  \\
Diff                        & 6.8/\textbf{6.2}  & 13.6/\textbf{12.5}   & 8.7/\textbf{6.1}   \\
 \midrule
OntoNotes                   & 62.6/\textbf{62.7} & 62.5/\textbf{62.9}   & 63.3/\textbf{63.4}  \\
\bottomrule
\end{tabular}
\caption{F1 on OntoNotes and WinoBias test set. WinoBias results have Type-1 and Type-2 in pro and anti stereotypical conditions. Average (Avg) and difference (Diff) of anti and pro stereotypical scores are shown.}
\label{tbl:winobias}
\end{table}

In \autoref{tbl:weat_wat}, we show the WEAT bias effects for cosine similarity and correlation on WAT dataset using the Pearson correlation coefficient.
We see that the proposed method can significantly debias for various biases in all word embeddings in both WEAT and WAT.
Especially in Word2Vec and fastText, almost all biases are debiased.

\autoref{tbl:sembias} shows the percentages where a word-pair is correctly classified as Definition, Stereotype or None.
We see that our proposed method succesfully debiases word embeddings based on results on \textbf{Definition} and \textbf{Stereotype} in SemBias.
In addition, we see that the SemBias-subset can be debiased for Word2Vec and fastText.

\autoref{tbl:winobias} shows the performance on WinoBias for \textbf{Type 1} and \textbf{Type 2} in \textbf{pro} and \textbf{anti} stereotypical conditions.
In most settings, the diff is smaller for the debiased than the original word embeddings, which demonstrates the effectiveness of our proposed method.
From the results for Avg, we see that debiasing is achieved with almost no loss in performance.
In addition, the debiased scores on the OntoNotes are higher than the original scores for all word embeddings.

\subsection{Comparison with Existing Methods}


\begin{table}[t]
\centering
\scalebox{0.8}{
\begin{tabular}{lccccc}
\toprule
                                            & GloVe            & HD  & GN-GloVe     & GP-GloVe    &  Ours      \\ \midrule
T1                        & $0.89^\dag$      & $0.97^\dag$                &  $1.10^\dag$              & $1.24^\dag$                               & $\bf 0.74^\dag$   \\
T2                    & $1.25^\dag$      & $1.23^\dag$                &  $1.25^\dag$              & $1.31^\dag$                               & $\bf 1.22^\dag$           \\
T5                              & 0.49             & -0.40                      &  \textbf{0.00}            & 0.21                                      & 0.35           \\
T6                           & $1.22^\dag$      & \textbf{-0.11}             &  $1.13^\dag$              & $0.78^\dag$                               & $1.05^\dag$           \\
T7    & 1.19             & 1.23                       &  1.11                     & \textbf{1.01}                             & 1.03                 \\ \bottomrule
\end{tabular}}
\caption{WEAT bias effects for the cosine similarity on prior methods and proposed method. $\dag$ indicates bias effects that are insignificant at $\al < 0.01$. T* are aligned with those in \autoref{tbl:weat_wat}.}
\label{tbl:weat_existing}
\end{table}

We compare the proposed method against the existing debiasing methods~\cite{Tolga:NIPS:2016,Zhao:2018ab,kaneko-bollegala-2019-gender} mentioned in \autoref{sec:related} on WEAT, which contains different types of biases.
We debias Glove\footnote{\url{https://github.com/uclanlp/gn_glove}}, which is used in~\newcite{Zhao:2018ab}.
All word embeddings used in these experiments are the pre-trained word embeddings used in the existing debiasing methods. 
Words in evaluation sets T3, T4 and T8 are not covered by the input pre-trained embeddings and hence not considered in this evaluation.
From \autoref{tbl:weat_existing} we see that only the proposed method debiases all biases accurately.
T5 and T6 are the tests for gender bias; despite prior debiasing methods do well in those tasks, they are not able to address other types of biases.
Notably, we see that the proposed method can debias more accurately compared to previous methods that use word lists for gender debiasing, such as~\newcite{Tolga:NIPS:2016} in T5 and~\newcite{Zhao:2018ab} in T6.

\subsection{Dominant Gloss vs All Glosses}
\label{sec:gloss}



\begin{table}[t]
\centering
\small
\begin{tabular}{lccc}
\toprule
\multirow{2}{*}{Embeddings} & Word2Vec 		& GloVe 						& fastText \\
                           	& Dom/All    	&  Dom/All  					& Dom/All   \\ \midrule
definition 	      			& 83.4/\textbf{83.9}                & \textbf{83.9}/83.4            & 92.5/\textbf{93.2} \\
stereotype        			& 12.7/\textbf{12.3}                & 11.8/\textbf{11.4}            & 4.8/\textbf{4.3} \\
none              			& \textbf{3.9}/\textbf{3.9}         & \textbf{4.3}/5.2              & 2.7/\textbf{2.5} \\ \midrule
sub-definition    			& 55.0/\textbf{57.5}                & \textbf{67.5}/\textbf{67.5}   & 77.5/\textbf{85.0} \\
sub-stereotype    			& 35.0/\textbf{32.5}                & \textbf{27.5}/\textbf{27.5}   & 12.5/\textbf{10.0} \\
sub-none          			& \textbf{10.0}/\textbf{10.0}       & \textbf{5.0}/\textbf{5.0}     & 10.0/\textbf{5.0} \\
\bottomrule
\end{tabular}
\caption{Performance obtained when using only the dominant gloss (Dom) or all glosses (All) on SemBias.}
\label{tbl:gloss}
\end{table}

In \autoref{tbl:gloss}, we investigate the effect of using the dominant gloss (i.e. the gloss for the most frequent sense of the word) when creating $s(w)$ on SemBias benchmark as opposed to using all glosses (same as in \autoref{tbl:sembias}).
We see that debiasing using all glosses is more effective than using only the dominant gloss.

\subsection{Word Embedding Benchmarks}

\begin{table}[t]
\centering
\small
\begin{tabular}{lcccccc}
\toprule
\multirow{1}{*}{Embeddings} & Word2Vec & GloVe & fastText          \\
                            & Org/Deb       & Org/Deb        & Org/Deb    \\ \midrule
WS                          & \textbf{62.4}/60.3           & 60.6/\textbf{68.9}   & 64.4/\textbf{67.0}     \\
SIMLEX                      & 44.7/\textbf{46.5}  & 39.5/\textbf{45.1}   & 44.2/\textbf{47.3}     \\
RG                          & 75.4/\textbf{77.9}  & 68.1/\textbf{74.1}   & 75.0/\textbf{79.6}     \\
MTurk                       & 63.1/\textbf{63.6}  & 62.7/\textbf{69.4}   & 67.2/\textbf{69.9}     \\
RW                          & 75.4/\textbf{77.9}  & 68.1/\textbf{74.1}   & 75.0/\textbf{79.6}     \\
MEN                         & 68.1/\textbf{69.4}  & 67.7/\textbf{76.7}   & 67.6/\textbf{71.8}     \\
\hline
MSR                         & \textbf{73.6}/72.6           & 73.8/\textbf{75.1}   & \textbf{83.9}/80.5              \\
Google                      & \textbf{74.0}/73.7           & 76.8/\textbf{77.3}   & \textbf{87.1}/85.7              \\ 
\bottomrule
\end{tabular}
\caption{The Spearman correlation coefficients between human ratings and cosine similarity scores computed using word embeddings for the word pairs in semantic similarity benchmarks.}
\label{tbl:sem}
\end{table}

\begin{figure*}[t]
\centering
    \begin{subfigure}[b]{0.45\textwidth}
        \centering
        \includegraphics[width=\linewidth]{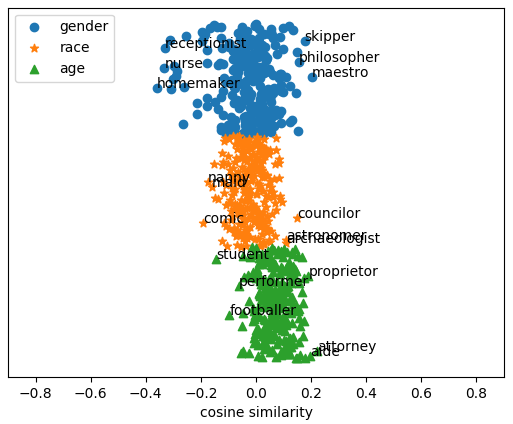}
        \caption{Original Word2Vec}
        \label{fig:bert_fuse}
    \end{subfigure}
    \hfill
    \begin{subfigure}[b]{0.45\textwidth}
        \centering
        \includegraphics[width=\linewidth]{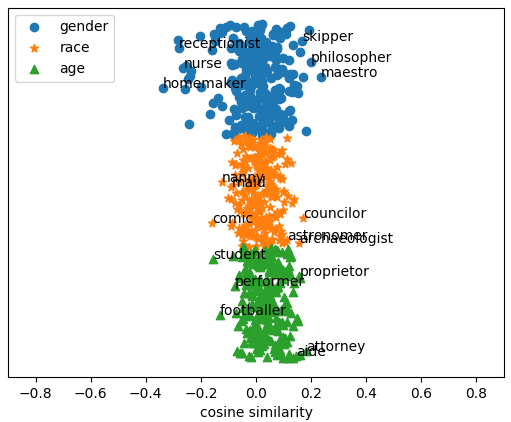}
        \caption{Debiased Word2vec}
        \label{fig:ft_bert_fuse}
    \end{subfigure}\hfill
    \caption{Cosine similarity between neutral occupation words for vector directions on gender ($\vv{he}-\vv{she}$), race ($\vv{Caucasoid}-\vv{Negroid}$), and age ($\vv{elder}-\vv{youth}$) vectors.}\label{fig:vizualization}
    \label{fig:visualization}
\end{figure*}

It is important that a debiasing method removes only discriminatory biases and preserves semantic information in the original word embeddings.
If the debiasing method removes more information than necessary from the original word embeddings, performance will drop when those debiased embeddings are used in NLP applications.
Therefore, to evaluate the semantic information preserved after debiasing, we use semantic similarity and word analogy benchmarks as described next.

\paragraph{Semantic Similarity:} 
The semantic similarity between two words is calculated as the cosine similarity between their word embeddings and
compared against the human ratings using the Spearman correlation coefficient.
The following datasets are used: Word Similarity 353~\cite[{\bf WS};][]{Finkelstein:2001}, {\bf SimLex}~\cite{J15-4004}, Rubenstein-Goodenough~\cite[{\bf RG};][]{Rubenstein:1965}, {\bf MTurk}~\cite{Halawi:2012}, rare words~\cite[{\bf RW};][]{W13-3512} and {\bf MEN}~\cite{P12-1015}.

\paragraph{Word Analogy:} 
In word analogy, we predict $d$ that completes the proportional analogy ``$a$ is to $b$ as $c$ is to what?'', for four words $a$, $b$, $c$ and $d$.
 We use CosAdd~\cite{Levy:CoNLL:2014}, which determines $d$ by maximising the cosine similarity between the two vectors ($\vec{b} - \vec{a} + \vec{c}$) and $\vec{d}$.
 Following \newcite{Zhao:2018ab}, we evaluate on {\bf MSR}~\cite{mikolov-yih-zweig:2013:NAACL-HLT} and {\bf Google} analogy datasets~\cite{NIPS2013_5021} as shown in \autoref{tbl:sem}.


From \autoref{tbl:sem} we see that for all word embeddings, debiased using the proposed method accurately preserves the semantic information in the original embeddings. 
In fact, except for Word2Vec embeddings on WS dataset, we see that the accuracy of the embeddings have \emph{improved} after the debiasing process, which is a desirable side-effect.
We believe this is due to the fact that the information in the dictionary definitions is used during the debiasing process.
Overall, our proposed method removes unfair biases, while retaining (and sometimes further improving) the semantic information contained in the original word embeddings.

We also see that for GloVe embeddings the performance has improved after debiasing whereas for Word2Vec and fastText embeddings the opposite is true.
 Similar drop in performance in word analogy tasks have been reported in prior work~\cite{Zhao:2018ab}.
 Besides CosAdd there are multiple alternative methods proposed for solving analogies using pre-trained word embeddings such as CosMult, PairDiff and supervised operators~\cite{Bollegala:IJCAI:2015,Bollegala:AAAI:2015,hakami-etal-2018-pairdiff}.
 Moreover, there have been concerns raised about the protocols used in prior work evaluating word embeddings on word analogy tasks and the correlation with downstream tasks~\cite{Natalie:NAACL:2018}.
 Therefore, we defer further investigation in this behaviour to future work.

\subsection{Visualising the Outcome of Debiasing}

We analyse the effect of debiasing by calculating the cosine similarity between neutral occupational words and gender ($\vv{he}-\vv{she}$), race ($\vv{Caucasoid}-\vv{Negroid}$) and age ($\vv{elder}-\vv{youth}$) directions.
The neutral occupational words list is based on \newcite{Tolga:NIPS:2016} and is listed in the Supplementary.
\autoref{fig:visualization} shows the visualisation result for Word2Vec.
We see that original Word2Vec shows some gender words are especially away from the origin (0.0).
Moreover, age-related words have an overall bias towards ``elder''.
Our debiased Word2Vec gathers vectors around the origin compared to the original Word2Vec for all gender, race and age vectors.

On the other hand, there are multiple words with high cosine similarity with the female gender after debiasing.
We speculate that in rare cases their definition sentences contain biases.
For example, in the WordNet the definitions for ``homemaker'' and ``nurse'' include gender-oriented words such as ``a wife who manages a household while her husband earns the family income'' and ``a woman who is the custodian of children.''
It remains an interesting future challenge to remove biases from dictionaries when using for debiasing.
Therefore, it is necessary to pay attention to biases included in the definition sentences when performing debiasing using dictionaries.
Combining definitions from multiple dictionaries could potentially help to mitigate biases coming from a single dictionary.
Another future research direction is to evaluate the proposed method for languages other than English using multilingual dictionaries.

\section{Conclusion}

We proposed a method to remove biases from pre-trained word embeddings using dictionaries, without requiring pre-defined word lists.
The experimental results on a series of benchmark datasets show that the proposed method can remove unfair biases, while retaining useful semantic information encoded in pre-trained word embeddings.

\bibliography{eacl2021}
\bibliographystyle{acl_natbib}

\end{document}


\maketitle

\section{Additional Description of Experiments}
We utilized a GeForce GTX 1080 Ti.
The debiasing is completed in less than an hour because our method is only a fine-tuning technique.
The parameter size of our debiasing model is 270,900.

\section{Evaluation Sources}
We used a bias comparing code\footnote{\url{https://github.com/hljames/compare-embedding-bias}} to evaluate \textbf{WEAT} dataset.
Since the \textbf{WAT} code was not published, we contacted the author to obtain the code and used it for evaluation.
We used the evaluation code from GP-GloVe\footnote{\url{https://github.com/kanekomasahiro/gp_debias}} to evaluate \textbf{SemBias} dataset.
We used AllenNLP\footnote{\url{https://github.com/allenai/allennlp}} to evaluate \textbf{WinoBias} and \textbf{OntoNotes} datasets.
We used \textit{evaluate\_word\_pairs} function and \textit{evaluate\_word\_analogies} in gensim\footnote{\url{https://github.com/RaRe-Technologies/gensim}} to evaluate \textbf{word embedding benchmarks}.

\section{Hyperparameters}

During optimisation, we used dropout~\cite{Srivastava2014DropoutAS} with probability $0.05$ to $\vec{w}$ and $E(w)$.
We used Adam~\cite{Kingma:ICLR:2015} with initial learning rate set to $0.0002$ as the optimiser to find the parameters $\vec{\theta}_{e}, \vec{\theta}_{c}$, and $\vec{\theta}_{d}$ and a mini-batch size of 4.
The optimal values of all hyperparameters are found by mimising the total loss over the development dataset
following a Monte-Carlo search.
We find these optimal hyperparameter values as $\alpha = 0.99998$,  $\beta = 0.00001$ and $\gamma = 0.00001$.
Note that the scale of different losses are different and the absolute values of hyperparameters do \emph{not} indicate the significance of a component loss. 
For example, if we rescale all losses to the same range then we have 
$L_c = 0.005\alpha$, $L_d = 0.269\beta$ and $L_a = 21.1999 \gamma$. 
Therefore, debiasing ($L_d$) and orthogonalisation  ($L_a$) contributions are significant. 

\section{Evaluation on Word Analogy Datasets}

\begin{table}[t]
\centering
\small
\begin{tabular}{lcccccc}
\toprule
\multirow{1}{*}{Embeddings} & Word2Vec & GloVe & fastText          \\
                            & Org/Deb       & Org/Deb        & Org/Deb    \\ \midrule
MSR                         & \textbf{73.6}/72.6           & 73.8/\textbf{75.1}   & \textbf{83.9}/80.5              \\
Google                      & \textbf{74.0}/73.7           & 76.8/\textbf{77.3}   & \textbf{87.1}/85.7              \\ \bottomrule
\end{tabular}
\caption{Accuracy for solving word analogies.}
\label{tbl:analogy}
\end{table}

 In word analogy, we predict $d$ that completes the proportional analogy ``$a$ is to $b$ as $c$ is to what?'', for four words $a$, $b$, $c$ and $d$.
 We use CosAdd~\cite{Levy:CoNLL:2014}, which determines $d$ by maximising the cosine similarity between the two vectors ($\vec{b} - \vec{a} + \vec{c}$) and $\vec{d}$.
 Following \newcite{Zhao:2018ab}, we evaluate on {\bf MSR}~\cite{mikolov-yih-zweig:2013:NAACL-HLT} and {\bf Google} analogy datasets~\cite{NIPS2013_5021} as shown in \autoref{tbl:analogy}.
 
 From \autoref{tbl:analogy} we see that for GloVe embeddings the performance has improved after debiasing whereas for Word2Vec and fastText embeddings the opposite is true.
 Similar drop in performance in word analogy tasks have been reported in prior work~\cite{Zhao:2018ab}.
 Besides CosAdd there are multiple alternative methods proposed for solving analogies using pre-trained word embeddings such as CosMult, PairDiff and supervised operators~\cite{Bollegala:IJCAI:2015,Bollegala:AAAI:2015,hakami-etal-2018-pairdiff}.
 Moreover, there have been concerns raised about the protocols used in prior work evaluating word embeddings on word analogy tasks and the correlation with downstream tasks~\cite{Natalie:NAACL:2018}.
 Therefore, we defer further investigation in this behaviour to future work.

\section{Occupation words}
We use following occupation words in analysis of visualization:

\noindent
accountant, acquaintance, administrator, adventurer, advocate, aide, ambassador, analyst, anthropologist, archaeologist, archbishop, architect, artist, artiste, assassin, astronaut, astronomer, athlete, attorney, author, baker, ballplayer, banker, barrister, bartender, biologist, bodyguard, bookkeeper, boss, boxer, broadcaster, broker, bureaucrat, butcher, cabbie, campaigner, captain, cardiologist, caretaker, carpenter, cartoonist, cellist, chancellor, chaplain, chef, chemist, choreographer, cinematographer, cleric, clerk, coach, collector, colonel, columnist, comedian, comic, commander, commentator, commissioner, composer, conductor, confesses, constable, consultant, cop, correspondent, councilor, counselor, critic, crooner, crusader, curator, custodian, dancer, dean, dentist, deputy, dermatologist, detective, diplomat, director, doctor, drummer, economist, editor, educator, electrician, employee, entertainer, entrepreneur, environmentalist, envoy, epidemiologist, evangelist, farmer, filmmaker, financier, firebrand, firefighter, footballer, gangster, gardener, geologist, goalkeeper, guitarist, hairdresser, headmaster, historian, homemaker, hooker, housekeeper, illustrator, industrialist, infielder, inspector, instructor, inventor, investigator, janitor, jeweler, journalist, judge, jurist, laborer, landlord, lawmaker, lawyer, lecturer, legislator, librarian, lieutenant, lifeguard, lyricist, maestro, magician, magistrate, maid, manager, marshal, mathematician, mechanic, mediator, medic, minister, missionary, mobster, musician, nanny, narrator, naturalist, negotiator, neurologist, neurosurgeon, novelist, nurse, observer, officer, organist, painter, paralegal, parishioner, parliamentarian, pastor, pathologist, pediatrician, performer, pharmacist, philanthropist, philosopher, photographer, photojournalist, physician, physicist, pianist, planner, playwright, plumber, poet, politician, pollster, preacher, president, principal, prisoner, professor, programmer, promoter, proprietor, prosecutor, protagonist, protege, protester, provost, psychiatrist, psychologist, publicist, pundit, radiologist, ranger, realtor, receptionist, researcher, restaurateur, sailor, saint, saxophonist, scholar, scientist, screenwriter, sculptor, secretary, senator, sergeant, servant, shopkeeper, singer, skipper, socialite, sociologist, soldier, solicitor, soloist, sportswriter, steward, stockbroker, strategist, student, stylist, substitute, superintendent, surgeon, surveyor, swimmer, teacher, technician, therapist, trader, treasurer, trooper, trucker, trumpeter, tutor, tycoon, undersecretary, understudy, valedictorian, violinist, vocalist, warden, warrior, welder, worker, wrestler, writer


\bibliography{anthology,eacl2021}
\bibliographystyle{acl_natbib}